\documentclass[letterpaper, 10 pt, conference]{ieeeconf}  

\IEEEoverridecommandlockouts                              

\usepackage{amsmath}
\usepackage{mathtools} 
\usepackage{graphicx}
\usepackage{float}
\usepackage{hyperref}  
\usepackage[ruled,vlined]{algorithm2e} 
\usepackage{color}

\usepackage{amssymb}
\newcommand{\R}{\mathbb{R}}
\newcommand{\V}{\mathcal{V}}
\newcommand{\E}{\mathcal{E}}
\newcommand{\Lap}{\mathcal{L}}

\newcommand{\norm}[1]{\left\lVert#1\right\rVert}
\DeclareMathOperator*{\argmin}{arg\,min}

\title{\LARGE \bf
Distributed Control of Truss Robots Using\\ Consensus Alternating Direction Method of Multipliers
}

\author{Nathan S. Usevitch$^{1}$, Trevor Halsted$^{1}$, Zachary M. Hammond$^{1}$, Allison M. Okamura$^{1}$, Mac Schwager$^{2}$
\thanks{*This work was supported by US National Science Foundation grant 1925030.}
\thanks{$^{1}$N.\ Usevitch, T.\ Halsted, Z.\ M.\ Hammond, and A.\ M.\ Okamura are with the Department of Mechanical Engineering, 
        Stanford University, Stanford, CA, 94305, USA
        {\tt\small usevitch@alumni.stanford.edu}}%
\thanks{$^{2}$M.\ Schwager with the Department of Aeronautics and Astronautice, Stanford University,
        Stanford, CA, 94305, USA
        {\tt\small schwager@stanford.edu}}%
}

\begin{document}

\maketitle
\thispagestyle{empty}
\pagestyle{empty}

\begin{abstract}

Truss robots, or robots that consist of extensible links connected at universal joints, are often designed with modular physical components but require centralized control techniques. This paper presents a distributed control technique for truss robots.  The truss robot is viewed as a collective, where each individual node of the robot is capable of measuring the lengths of the neighboring edges, communicating with a subset of the other nodes, and computing and executing its own control actions with its connected edges. Through an iterative distributed optimization, the individual members utilize local information to converge on a global estimate of the robot's state, and then coordinate their planned motion to achieve desired global behavior. This distributed optimization is based on a consensus alternating direction method of multipliers framework. This distributed algorithm is then adapted to control an isoperimetric truss robot, and the distributed algorithm is used in an experimental demonstration. The demonstration allows a user to broadcast commands to a single node of the robot, which then ensures the coordinated motion of all other nodes to achieve the desired global motion.  

\end{abstract}


\section{Introduction}
A longstanding challenge in robotics is designing a single robotic system that is capable of performing a variety of tasks and operating in a variety of different environments. One approach to enabling this increased flexibility is to create robots that are capable of changing their overall shape to respond to different tasks or environments. The potential for this active shape change has been a key driver in the development of truss-like robots that consist of a set of length-changing edges interconnected by a number of universal joints to create a truss- or mesh-like structure.  These robots have been proposed for applications in which a high-degree of shape flexibility is required, such as exploring planets \cite{hamlin1996tetrobot, curtis2007tetrahedral}, or shoring up rubble in disaster zones \cite{spinos2017towards}. 

\begin{figure}[tb] 
\centering \includegraphics[width=\columnwidth]{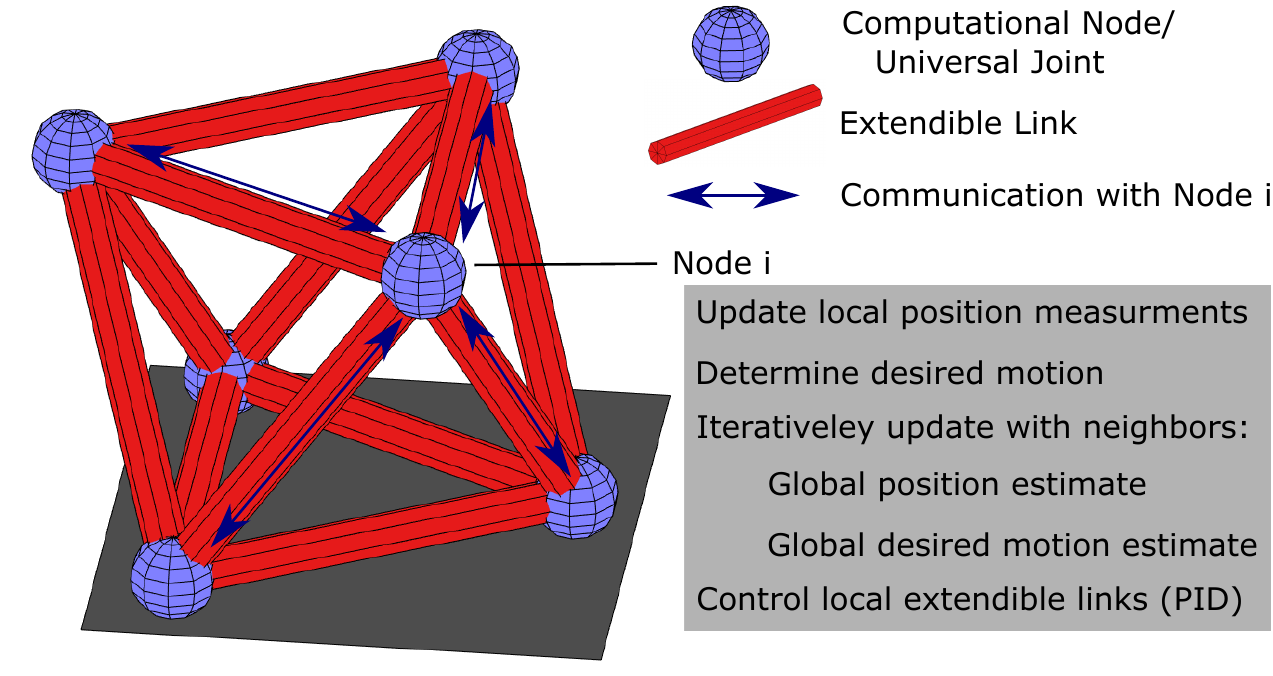}
\caption[Distributed Control Schematic]{A schematic of a truss robot, and an algorithmic sketch of the distributed algorithm operating at one of the computational nodes. The truss robot consists of extensible links connected at universal joints. Under distributed control, each node takes local measurements and performs iterative communication with the neighboring nodes to determine the state of the robot and the local control action that it needs to take. }
\label{fig:D_Control_Schematic}

\end{figure}

Modularity has been frequently cited as one of the important advantages of truss robots \cite{hamlin1996tetrobot,curtis2007tetrahedral,yu2008morpho}. In this paper, we consider truss robots as a collective of many individual members, each with their own sensing, computation, and actuation, who coordinate their motion to achieve desirable overall results. Research on robotic collectives or swarms often draws inspiration from biological collectives such as swarms of fish, birds, and insects, in which each member of the collective is capable of individual motion \cite{mohan2009extensive}. However, another type of collective exists in which individual members of the swarm are physically connected into a structure, such as when colonies of ants combine to form structures such as bridges or nests, or when slime mold organisms aggregate and collectively locomote \cite{malley_Eciton_robotica,mondada2005PhysicalInteractionincollectiverobots,umedachi2010decentralized_Slime_Mold}. In a swarm where every agent is physically disconnected, each member must be capable of moving on its own, and the motion of each component is typically not directly altered by the motion of its neighbors. In a physically interconnected collective, the shared connections impose constraints on each member's motion, and the motion of one member directly changes the position of the other members throughout the collective. The coordinated control of a physically interconnected collective thus poses additional challenges while also allowing for a collective to achieve interesting behaviors --- even when the individual members are capable of only simple behaviors. 

In this paper, we consider truss robots as physically interconnected collectives.  We define the nodes of the system as the universal joints between edges, and assume that each node is capable of computation and communication with the other nodes to which it is connected with a physical edge. We present distributed algorithms that allow each node to determine the shape of the overall robot and coordinate their motions to minimize a cost function and achieve desired motions, even if the desired motions are only known to a subset of the nodes. We first define the problem and provide an outline of our algorithm. We then present the mathematical underpinnings of both the state estimation and control components of the algorithm, which are based on a consensus formulation of alternating direction method of multipliers (ADMM) with the ability to locally enforce constraints. We apply these algorithms to distributed state estimation for truss robots using local measurements at each node. Next, we use the same ADMM framework to determine which control actions to apply in order to achieve desired motion objectives.  We demonstrate both state estimation and control in simulation. We then discuss the adaptation of this framework to controlling an isoperimetric robot, a type of robotic truss that includes unique constraints that must be added to the control and estimation algorithm. The isoperimetric robot is then used in a demonstration in which a user can teleoperate one node of the robotic truss while the robot uses the distributed algorithm to determine the proper individual contributions of each member of the collective. 

The contributions of this paper are (1) Application of a distributed algorithm based on the alternating direction method of multipliers to distributed state estimation and control of (2) Demonstration of this algorithmic approach using an isoperimetric robot.

\subsection{Related Work}

This paper builds on past work on distributed control of truss robots. A key contribution in this area is the TETROBOT project, which developed a set of modular control algorithms that work in conjunction with modular hardware \cite{hamlin1997tetrobot}. The distributed algorithms presented in \cite{hamlin1997tetrobot} divide the nodes of the truss into two categories: controlled and unconstrained. The unconstrained nodes move to minimize a cost function, and the controlled nodes have a specified trajectory. The distributed algorithms use the chain-like kinematic architecture of the robot to coordinate motions for each actuator, and also allow the algorithms to account for dynamic effects \cite{lee2002dynamic}. However, these algorithms only allow for specification of individual node motion. For example, in the controller of each node it is impossible to control the motion of the center of mass, which is useful in enabling locomotion. In \cite{usevitch2017linear, usevitchTRO_LAR}, the kinematics of arbitrary networks of extensible links are described, and centralized, nonlinear optimization techniques are used to generate trajectories for these robots, without discussion into how to adapt these centralized methods to decentralized control.  

Our work also builds on past work on distributed optimization. Our algorithms are based on a consensus formulation of the alternating direction method of multipliers (ADMM), which is discussed in detail in \cite{boyd2011ADMMOverview}. The ADMM framework allows for the distributed solution of optimization problems. It is extended to a multi-agent distributed computation framework in \cite{chang2014inexactconsensus_ADMM,mateos2010distributedsparelinreg}. Consensus ADMM has also been used for multi-target tracking \cite{shorinwa2020distributed_multitarget}. In this work, we adapt consensus ADMM to include the handling of linear constraints known to only a subset of the nodes, and apply these results to both distributed estimation and control.

Past work has also focused on the physical construction of truss robots.  Many physical embodiments of truss robots have been demonstrated, with different types of linear actuators serving as the edges, and with different mechanical designs for the universal joints. Tensegrity robots, a related family of robots consisting of a network of rigid compression elements suspended in a network of compliant cables, have been studied for applications as landers and rovers on other planets, and have demonstrated locomotion and other abilities \cite{paul2005gait,friesen2014ductt,bruce2014superball,sabelhaus2015sys_designSUPERball}. While tensegrity robots look similar to the truss robots we consider here, their kinematics, dynamics, and control are quite different.  The methods we develop here might be adaptable to control of tensegrity robot, but we do not consider such extensions in this paper.  A new soft architecture for truss robots, called the isoperimetric robot, was recently introduced in \cite{usevitch2020untethered}. The algorithmic approach we describe in this paper is valid for soft isoperimetric truss robots, as well as traditional truss robot designs.  We demonstrate the application of our algorithms in a physical experiment using an isoperimetric system similar to the one presented in \cite{usevitch2020untethered}.

\section{Problem Formulation and Algorithmic Sketch} \label{Sec:Formulation}

 The truss robot is defined as a framework that consists of a graph $G$ and vertex positions $p_i \in \R^d$.  The graph is denoted as $G=( \V, \E )$, where $\V= \left\{ 1, \ldots, n \right\}$ is the set of vertices and $\E=\left\{ \ldots, \{i,j\}, \ldots \right\} $ is the set of undirected edges. The geometry of the robot is fully represented by the concatenation of all vertex positions $x=[p_1^T, p_2^T,..., p_n^T]^T$. We define a length vector $L$, which contains the lengths of all edges in the graph such that the $k$th element is


\begin{equation}
L_{k}=\|p_i-p_j\| \quad \forall \left\{ i,j \right\} \in \E.  
\label{eq:D_Norms}
\end{equation}
The motions of the edges and the vertex are related through the expression

\begin{equation}
    \dot{L}=R(x) \dot{x},
\end{equation}
\noindent where $R(x)$ is the scaled rigidity matrix further analyzed in \cite{usevitch2017linear}.

We suppose that a computation, communication, and sensing unit, which we call a node, is located at each vertex of the graph.  We present distributed techniques that allow each node to determine the shape of the overall robot, as well as methods to coordinate the control of the edge lengths to achieved desired motion of the nodes. Our algorithms also allow a high-level planner, or even a human operator, to send commands in task space.  The robot then coordinates control actions with its neighbors to achieve this command.  A schematic of this control architecture with the physical connections between nodes, node-to-node communication, and the opportunity for communication from an offboard source, is shown in Fig.~\ref{fig:D_Control_Schematic}. 

Algorithm 1 gives the overall structure of our algorithm. During each control loop, each node first acquires measurements from its local sensors. The form of these measurements will be discussed in Sec.~\ref{Sec:D_StateEstimation}, and may include the edge lengths of the edges adjacent to the node, or the relative positions of neighboring nodes. This information is used to complete an iterative ADMM optimization, where estimates of the robot state are iteratively communicated with the neighboring nodes and then updated for a fixed number of iterations.  The result of this step is that each node converges to a shared estimate of the robot's state in a global frame. Each node then uses this state information, as well as other knowledge it may have about constraints on its motion or the robot's motion, to perform another iterative ADMM optimization to compute the optimal motion of each node. At this point, the messages exchanged between neighboring nodes are estimates of the velocities of all nodes in the network, and no information is exchanged about which constraints a certain node may be trying to satisfy. After converging to the optimal solution for how all nodes should move, the node motions are translated into actuator commands using both the estimate of the robot's state and the desired motion of the nodes. The corresponding commands are sent to the physical actuators, and the process repeats. In the following sections, we will discuss each component algorithm in detail.  

\begin{algorithm}[t]
\SetAlgoLined
 $\hat{x}^1 \gets Initial Guess at Robot State$\;
 \While{1}{
  $M \leftarrow Acquire Measurements() $\;
  $\hat{x} \leftarrow Compute State Estimate (M, \hat{x}^{k-1})$\;
  $\dot{x} \leftarrow Coordinate Motion(\hat{x},Local Constraints)$\;
  $\dot{L} \leftarrow Compute Action(\hat{x},\dot{x})$\;
  $Apply Control(\dot{L})$
  }
 \caption{Distributed Truss Control}
\end{algorithm}

\section{Consensus ADMM Framework}
Both the distributed state estimation and control algorithms are based on a consensus ADMM framework. This section  introduces this framework generally, and the following sections apply it to the specific problems of state estimation and control of truss robots. We define the general, centralized problem that we are are solving throughout this paper as  
\begin{align}  
  \min_{x} \quad & J(x) = \sum_{i=1}^n J_i(x)   \label{eq:D_cent_J} \\
   \text{subject to} \quad & Ax=b  \label{eq:D_cent_Axb} 
\end{align}

We distribute this problem across $n$ computational nodes, with the set of nodes the neighbor node $i$ defined as $N_i$. Each node maintains its own, local copy of the total state vector $x_i$. We divide the cost function into local components $J_i(x)$ such that $\sum_{i=1}^n J_i(x)= J(x)$, and each node maintains a local copy of a subset of the linear constraints $A_i x_i = b_i $. Satisfaction of all of the local constraints must ensure satisfaction of the constraints to the centralized problem such that if $x$ satisfies $A_i x= b_i$ for all $i$, then $x$ satisfies $Ax=b$.

Each node then solves the following optimization problem

\begin{align}  
   \min_{x_i} \quad & J_i(x_i) \label{eq:D_cost_G} \\ 
   \text{subject to} \quad &  A_i x_i = b_i \\  
   & x_i=x_j, \quad \forall j \in N_i. \label{eq:D_consist}
\end{align}

Equation~\ref{eq:D_consist} is a consistency constraint, which requires that the copy of the state vector at node $i$ must equal the state vector at all neighboring nodes. Solving this distributed problem then yields a solution that is equivalent to the solution to the centralized problem in Eqs.~\ref{eq:D_cent_J} and ~\ref{eq:D_cent_Axb}.  To solve these coupled optimization problems, we form the augmented Lagrangian

\begin{align}
    \begin{split}
        \Lap = \sum_{i\in V} \bigg ( & J_i(x_i)+ r_i^T(A_i-b) + \frac{\alpha_r}{2} \norm{A_ix_i-b_i}^2 +  \\ & \sum_{j\in N} [ \lambda_{ij}^T(g_{ij}-x_i) + \nu_{ij}^T(g_{ji}-x_j) ]  +  \\ &  \frac{\alpha_p}{2}\sum_{j \in N_i}\big (\norm{g_{i,j}-x_i}^2+ \norm{g_{ij}-x_j}^2 \big )  \bigg ),
    \end{split}
\end{align}

\noindent where $g_{ij}$ are auxiliary primal variables that encode the consistency constraints,  $r_i$ are the Lagrange multipliers associated with the linear constraints, and $\lambda$ and $\nu$ are the Lagrange multipliers associated with the consistency constraints. The hyperparameters $\alpha_r$ and $\alpha_p$  tune the sensitivity to disagreement between the neighbor's estimates and the violation of the local linear constraints.  We then iteratively update each set of variables in the augmented Lagrangian through a gradient ascent step on the Lagrange multiplier variables ($\nu$, $\lambda$, and $r$) and a minimization step of the primal variables ($x$ and $g$) variables as follows: 

\begin{align}
    &\lambda_{ij}^{(k+1)}= \lambda_{ij}^{(k)}+\alpha_p\big (g_{ij}^{(k)}-x_i^{(k)}\big ) \quad \forall(i,j)\in \E  \label{eq:D_update_lamb} \\  
    &\nu_{ij}^{(k+1)}= \nu_{ij}^{(k)}+\alpha_p\big (g_{ij}^{(k)}-x_j^{(k)}\big ) \quad \forall(i,j)\in \E \\
    &r_{ij}^{(k+1)}= r_{ij}^{(k)}+\alpha_r \big (A_i x_i^{(k)} - b_i\big ) \\
    &x^{(k+1)} = \argmin_x \{\Lap(x,r^{(k+1)},g^{(k)},\lambda^{(k+1)}_{ij},\nu_{ij}^{(k+1)} \}  \label{eq:D_update_1} \\
    &g^{(k+1)}= \argmin_g\{\Lap(x^{k+1}, r^{(k+1)}, g, \lambda^{(k+1)}_{ij},\nu_{ij}^{(k+1)} \}  \label{eq:D_update_g}
\end{align}

The update in Eq.~\ref{eq:D_update_1} can be solved exactly in a distributed manner for each $x_i$ because the augmented Lagrangian is separable. As shown in \cite{mateos2010distributedsparelinreg, chang2014inexactconsensus_ADMM}, substituting $p_i=\sum_{j \in N_i}\lambda_{ij}+\nu_{ij}$ and assuming the initialization $p_i^{(0)}=0$ causes Eq.~\ref{eq:D_update_g} to become
\begin{equation}
    g_{ij}=\frac{1}{2}(x_i+x_j).
\end{equation}
Using this expression we can rewrite the iterative steps Eqs.~\ref{eq:D_update_lamb}-\ref{eq:D_update_g}  in a manner in which each agent can compute its updates in parallel as follows:

\begin{align}
& p_i^{k+1}=p_i^{k}+\alpha_p \sum_{j\in N_i} (x_i^k-x_j^k), \\
& r_i^{k+1}=r_i^{k}+\alpha_r(A_i x_i^k -b_i),  \label{eq:D_pupdates} \\
\begin{split}
    & x_i^{k+1}=\argmin_{x_i}\bigg (J_i(x_i) + (p_i^{k+1})^T x_i +  (r_i^{k+1})^T(A_i x^k_i - b_i) + \\ & \quad \alpha_p \sum_{j\in N_i} \norm{ x_i-\frac{x_i^{k}+x_j^{k}}{2} }_2^2 + \alpha_r \norm{A_i x_i - b_i }_2^2 \bigg ).  \label{eq:D_argmin_update}
\end{split}
\end{align}

Each node iteratively perform these updates, which require each node to communicate their estimates of the state with the neighbors and solve the optimization problem in Eq.~\ref{eq:D_argmin_update}. By performing these update steps iteratively, each agent only communicates with its neighbors, and their estimates  converge to a shared estimate that satisfies the local constraints. If $J(x)$ is convex, then $x_i^{(k)}$ will approach the optimal centralized solution $x^*$ as the number of iterations increases to infinity. 

\subsection{Quadratic Cost Function}
A special case that is relevant for our work is where the cost function $J_i(x)$ is of the quadratic form

\begin{equation}
   J_i(x_i)= \| D_i x_i +f_i \|^2.
\end{equation}

This form allows us to compute the analytic solution to the optimization problem in Eq.~\ref{eq:D_argmin_update} as follows
\begin{align}
\begin{split}
x_i^{k+1}=M^{-1}\bigg( & 2\alpha_r A_i^T b_i -p_i^{k+1} - A_i^T r_i^{k+1} + \\ &\alpha_p \sum_{j\in N_i} (x_i^k + x_j^k) \bigg) ,  \label{eq:D_Quad_update}
\end{split}
\end{align}
where $M$ is given by
\begin{equation}
    M=D_i(x_i)^T D_i(x_i)+2 \alpha_r A_i^T A_i + 2 \alpha_p d_i,
\end{equation}
and $d_i$ is the degree of each node. We note that because the updates in Eqs.~\ref{eq:D_pupdates}-\ref{eq:D_argmin_update}  are performed iteratively, the matrix $M^{-1}$ is unchanged throughout the round of iterations for a given problem. This allows for very efficient computation, because the matrix inverse needs to be computed only once per problem, and then the updates performed at each iteration only require multiplication of precomputed matrices and changing state estimates. 

\subsection{Convergence Criteria}
Using the consensus ADMM approach to distributed optimization requires determining when to terminate the iterations. In a distributed setting, determining when a stopping criteria is reached is challenging because each node does not have all of the information. For example, if one node has converged and all constraints are satisfied, this does not guarantee that another node, elsewhere in the network, has converged on the proper solution. In addition, the convergence rate is also influenced by the selection of the hyperparameters $\alpha_p$ and $\alpha_r$. Throughout this paper we empirically select hyperparameters and run the optimization for a fixed number of iterations that have empirically been demonstrated to result in good convergence. The fact that the communication graph is set by the physical connections of the robot and does not change indicates that the convergence behavior during experiments will be similar to what would be observed on an actual robot.

\section{Distributed State Estimation} \label{Sec:D_StateEstimation}
We first consider the problem of state estimation, or the problem of determining enough about the state of the robot to allow each node to plan and compute control actions. The amount of information about the robot's state required depends on the amount of information needed to compute the local cost function $J_i(x)$, translate a planned motion into action, and evaluate any constraints. We select cost functions that only require each agent to know its neighbors' positions. Knowledge of positions and planned motions of the neighboring nodes is sufficient to translate the planned motion into commands for each actuator. 
In addition to minimizing a cost function, several different types of constraints have been discussed in the literature that are inherent with the physical construction of the robot. These constraints include (i) that each edge stay within a maximum and minimum length, (ii) that edges do not collide, (iii) that the angle between connected edges remain above a certain threshold, and (iv) that the robot avoids singular configurations, which is equivalent to avoiding configurations where the framework describing the robot is no longer infinitesimally rigid. The constraints (i) and (iii) on the edge lengths and angles, respectively, only require that each node have information about the neighboring nodes. The edge collision constraint (ii) requires that nodes are able to determine a region of potential collision, which could also be achieved with local information. In the general case the constraint (iv) to maintain infinitesimal rigidity requires that each node be aware of the location of nodes in the network that are not its immediate neighbor. For this reason, we develop a distributed estimation algorithm where each node reconstructs the entire state of the robot, and does so by only using local measurements and then communicating estimates of the robot's state with the neighboring nodes. We evaluate the case where the nodes are able to measure either the relative distance to the neighboring nodes, or the relative positions of the neighboring nodes. 

We also note that these algorithms are contingent on all of the nodes having aligned reference frames. In practice this can be achieved by equipping each node with an inertial measurement unit capable of measuring a gravity vector and a vector indicating magnetic north. From these two vectors, it is possible to reconstruct an aligned set of frames.

\subsection{State estimation from Relative Position Estimates}

We first determine the overall configuration of the robot by assuming that each node in the network is capable of computing estimates of the position of its neighbors in a local reference frame. In a conventional truss robot, this naturally occurs if each node has knowledge of the orientation and length of each incident edge. This measurement could also be achieved if each node can visually determine the distance and position of its neighbors. We express this relative position measurement as $v_{i,j}$ such that $p_i+v_{i,j}=p_j$. Combining all of these expressions we obtain the following expression which can be written as a summation over all of the nodes in the network 
\begin{equation}
    J(x)=\sum_i^n  \sum_{j \in N_i} \|p_j- p_i - v_{ij} \|^2= \sum_i^n J_i(x) \label{eq:D_J_position}
\end{equation}

This cost function is invariant to translation of the robot, which could create ambiguity on how the robot is moving over time. We resolve this by including additional linear constraints of the form $Ax=b$ on the position of the nodes. One option is to assume that the centroid of the robot is located at the origin, which is possible if we express the constraint as $1^T \otimes I_3 x =0$.  Alternatively, if we know the position of one anchor node, referred to as node $i$, we can express the constraint as $e_i^T \otimes I_3 x =p_i$, where $e_i$ is an $n\times 1$ vector of zeros where element $i$ is equal to $1$. 

Combining the cost function in (\ref{eq:D_J_position}) and the linear constraints, we formulate a distributed optimization problem of the form posed in (\ref{eq:D_cost_G})-(\ref{eq:D_consist}), which can be solved through the iterative updates in Eqs.~\ref{eq:D_pupdates}-\ref{eq:D_argmin_update}. This cost function is quadratic, meaning that the optimization problem in Eq.~\ref{eq:D_argmin_update} can be solved analytically using Eq.~\ref{eq:D_Quad_update}, which leads to very efficient computation. 

An important parameter is the number of relative position measurements necessary to reconstruct the robot shape.  Relative position measurements are sufficient to reconstruct any connected graph, and a connected graph must have at least $n-1$ edges. A robot must have at least $n-1$ edges in order to have a unique solution to determining the global positions based on relative position measurements. In practice, the truss robots are infinitesimally rigid, meaning that they have at least $3n-6$ edges in 3D and $2n-3$ edges in 2D, and thus there are redundant measurements if all relative positions are measured.  This strategy has the effect of using this redundant information to improve the position estimate. 

\subsection{State Estimation from Relative Distance Measurements}

Another option for reconstructing the global state of the robot is to assume that each node knows its distance to all of the neighboring nodes, but not their positions. This is achieved if each node knows the lengths of all the adjacent actuators. In this case, the cost function is expressed as 

\begin{equation}
    \sum_{i=1}^{N_L} \| L_i(x) - L_{m,i} \|^2 . \label{eq:D_J_measure}
\end{equation}
We divide this cost between the computational nodes by having each node compute the cost based on only the adjacent edges. This cost function is invariant to both translation and rotation.  We remove this invariance by defining at least $6$ linearly independent constraints on the positions of the nodes which we express as $Ax=b$. In 2D, we need only define $3$ linearly independent constraints. In practice, we define these constraints by defining the feet of the robot, or the nodes of the robot that form the support polygon of the robot, to be fixed in all dimensions.  In a physical system these nodes could detect that they were on the ground using contact sensing. 

Similar to the case of relative positions, we combine the cost function and constraints. The cost function in Eq.~\ref{eq:D_J_measure} is nonconvex, indicating that several local minima may exist. This cost function is exactly equivalent to reconstructing a graph based on its edge lengths, a key problem in rigidity theory that is discussed in some depth in  \cite{usevitchTRO_LAR}. Depending on the number of edges, different classes of solutions may exist. For truss robots, we consider only graphs that are minimally rigid or over-constrained because those are the only graphs where the node motion can be fully controlled by changing the edge lengths. If the graph is minimally rigid, there are $3n-6$ edges, which constitute the minimum number of edges to fix the position of the nodes. However, this also means that there is no redundant information that can be utilized to reduce the effect of noisy measurements. Additional edges in the graph that lead to the graph being over-constrained could allow for improved behavior under noisy measurements. We also note that each iteration of the update procedure with the distance objective function requires solving the nonlinear optimization problem using an iterative numeric solver. This leads to substantially slower performance than the analytic solution to the quadratic optimization problem presented in Eq.~\ref{eq:D_J_position}. 

\subsection{Comparison}

We have presented two algorithms for state estimation, one that utilizes relative position information to each neighbor, and another that utilizes relative distance information. A drawback of using relative position estimates is that it requires that more information be gathered in the measurements, but is also has several advantages: the optimization problem has a unique solution, redundant information is incorporated to reduce the effect of noisy measurements, and it is computationally efficient because the optimization that is part of each iteration can be solved analytically. The estimation scheme using relative distance leads to an optimization problem that could potentially have multiple solutions, and may not include any tolerance to noisy measurements. In Sec.~\ref{Sec:Sim_Results} we will use simulation to further compare these two approaches. 
A key point is that during the iterative state estimation routine, the nodes communicate only their estimates of the global state with their neighbors, and do not communicate their measurements directly. This potentially increases the generality of the algorithm, because other types of measurements could be incorporated in the cost function or constraints at each node, while the information exchanged between nodes remains the same. 

\section{Distributed Control Algorithm}

We now present a distributed algorithm that allows the nodes of the robot to determine how to coordinate the motion of the actuators to minimize a cost function while satisfying constraints. In the previous section, the nodes complete an ADMM optimization where the decision variable is the location of each node in a shared reference frame. For the case of control, the decision variable is a vector of the velocity of each of the nodes, $\dot{x}$. Controlling in the space of velocities provides a natural method to encode behavior and also simplifies the treatment of constraints. We express the physical feasibility constraints as a function of the position of the nodes, $f(x)<0$. While the constraints $f(x)<0$ are nonlinear, the derivative of these constraints is linear in the velocity of the nodes $\frac{df(x)}{dt}=\frac{\partial f(x)} {\partial x} \dot{x}$. Our approach is to compute which constraints are active, and then use our algorithm to enforce the linear constraint that the nodes do not move along the gradient direction of increasing constraint violation. A key advantage of the distributed approach is that constraint information can be held at each node, and none of the other nodes need to be aware of a constraint for it to be obeyed.  For example, if a user seeks to teleoperate the robot, velocity commands can be sent to a single node of the robot and used as a local linear constraint to ensure that the node satisfies the specified motion. No other node of the robot needs to know  the command. If a truss robot is moving autonomously through a cluttered environment and one node is close to colliding with an external obstacle, the node can enforce a constraint to stop moving in a given direction. Despite the fact that no other nodes are aware of the obstacle, the distributed control algorithm will ensure that no other node moves in a way that violates the constraint. In addition, constraints that involve many nodes will be obeyed even if they are sent to a subset of the nodes. For example, a constraint that the center of mass move with a particular velocity can be broadcast to one or more of the nodes, and the optimization will ensure that all nodes move in a way that satisfies the constraint. 

For the case of control, we consider two different cost functions. Each deals with a cost that is computed as a sum of costs for each actuator. We distribute this cost by having each node consider the cost of all adjacent actuators.
 
\begin{equation}
    J(\dot{x})=\|\dot{L}(x)\|= \| R(x)\dot{x} \|^2= \sum_{i=1}^n \|\dot{L}_{i,j\in N_i}(x)\| 
\end{equation}
The second objective seeks to minimize the deviance of the edges from some nominal edge length. This behavior can be expressed through the following cost function, which is a function of the position of the nodes

\begin{equation}
    \|L(x)- L_{nominal} \|^2  \label{eq:D_form_pos}
\end{equation}
 where $L_{nominal}$ is a vector of the nominal length of each edge. However, our control algorithm requires a command in terms of the velocities, and not positions. To translate this concept of maintaining nominal edge lengths to an objective that is a function of velocity, we use as the cost function the norm squared of the difference between the velocity and the gradient of Eq.~\ref{eq:D_form_pos}, 
 
 \begin{equation}
    \|\dot{x} - (L(x)-L_{nominal})^T R^T(x) \|^2.   \label{eq:D_cost_quad}
\end{equation}
 
In addition to the cost function, we also impose the constraints that the ground feet of the robot remain stationary as

\begin{equation}
C\dot{x}=0.
\end{equation}

We can also encode any other constraint that is linear in the velocity of nodes and is of the form 
\begin{equation}
    A \dot{x} = b.
\end{equation}
If we use the mass matrix for $A$, this allows us to control the center of mass of the robot. We can also define the $A$ matrix to specify the velocity of a certain node if it is of the form $[0, 0, \dots, I_3, \ldots, 0]$

To perform the control, we select either the cost function given in Eq.~\ref{eq:D_form_pos} or \ref{eq:D_cost_quad}, and a set of local constraints to define the following optimization problem:

\begin{align}  
   &\min_{\dot{x}_i}    
   \begin{aligned}[t]
        J_i(\dot{x}_i)  \label{Motion_Primitive}
   \end{aligned} \\
   &\text{subject to} \notag \\
   &\quad \begin{bmatrix} A_i \\ C_i \end{bmatrix} \dot{x}_i=\begin{bmatrix} b_i \\ 0 \end{bmatrix}  \label{eq:lin_control} \\
   &\quad  \dot{x}_i=\dot{x}_j, \quad \forall j \in N_i,   \label{eq:D_vel_con} 
\end{align}
where $\dot{x}_i$ is the estimate at node $i$ of the velocity of all nodes of the robot. 

\section{Simulation Results}
 We validate the algorithm via simulation in this section, and in hardware experiments in the next section. We first present results on state estimation, and then results on the control algorithms. 

\subsection{State Estimation} \label{Sec:Sim_Results}
\begin{figure*}[tb] 
\centering \includegraphics[width=\textwidth]{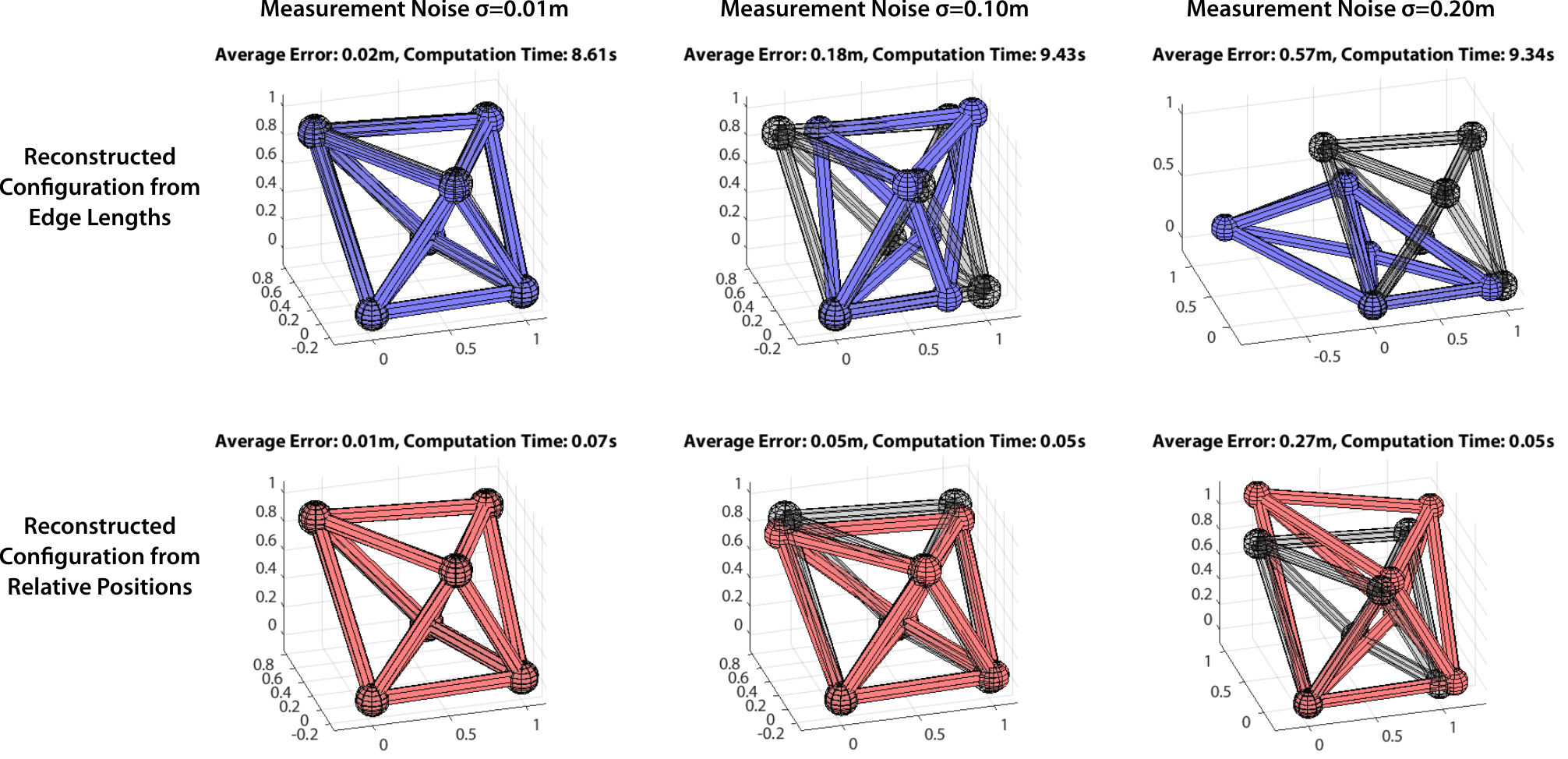}
\caption[Effect of Noise on State Estimation]{The reconstructed state with varying levels of noise injected into the measurements. The top row shows the state estimate when each node measures its relative distance to its neighbors. The bottom row shows the state estimate when each node measures the relative position of each of its neighbors. With increasing noise, the estimate using relative position information produces better estimates. In addition, the relative position information allows far more efficient computation.}
\label{fig:D_Noise_Rejection}

\end{figure*}

To examine the performance of the state estimation algorithms, we performed simulations using both the cost function in which each node measures the relative position of the neighboring nodes (Eq.~\ref{eq:D_J_position}), and the cost function where each agent measures the distance to the neighboring nodes (Eq.~\ref{eq:D_J_measure}). Fig.~\ref{fig:D_Noise_Rejection} shows the results for both techniques using several different levels of noise in the measurements of either the distances or relative positions. We choose an octahedral robot shape, and perturb it slightly from its nominal configuration by starting all edges at a length of 1 m, then increasing or decreasing the lengths based on a distance generated by a normal distribution with 0 mean and variance of 0.25 m. This allows us to demonstrate that the resulting convergence does not leverage the symmetry of a uniform graph. Both cost functions use linear constraints to fix the height of all three support feet, in addition to fixing the $x$ and $y$ position of one foot and the $y$ position of the third foot. We complete 200 rounds of the ADMM iterations to minimize the cost function and satisfy the constraints using the updates in Eqs.~\ref{eq:D_pupdates}-\ref{eq:D_argmin_update}. We perform three different trials, increasing the variance of normally distributed noise that we use for the measurements. For the relative position estimates we use the analytic solution shown in Eq.~\ref{eq:D_Quad_update}. When using the relative distance measurements, we use MATLAB's fminunc solver to solve the optimization in Eq.~\ref{eq:D_argmin_update}. To increase computation speed of the fminunc solver, we analytically compute the gradient of the objective function and provide it to the solver.  We perform all computation on a laptop computer (Intel Core i7 Processor, 4 cores, 2.80 GHz, 16GB RAM).

The top row of Fig.~\ref{fig:D_Noise_Rejection} shows the results based on relative distances, and the bottom row shows results for relative positions, while each column corresponds to a different level of noise injected into the measurements. For the case of low noise, both results converge to a solution that appears approximately identical to the nominal shape of the graph. To quantify the closeness of the fit, we compute the average error as the average distance between each node's estimated and true position. As the noise level increases, the average error of both estimation schemes increases. Overall, the error using the relative position measurements (Eq.~\ref{eq:D_J_position}) is lower than for relative distances (Eq.~\ref{eq:D_J_measure}). This is expected, because the relative position measurements contain more information than the relative distance measurements. Another key difference between these two optimizations is the computation time. Using the relative position measurements and the resulting quadratic cost function, the maximum duration of the 200 iterations was 0.071 seconds. Using the relative distance measurements requires completing the iterative optimization using fminunc every time step, and led to a maximum computation time of 9.43 seconds. 

\begin{figure}[tb] 
\centering \includegraphics[width=\columnwidth]{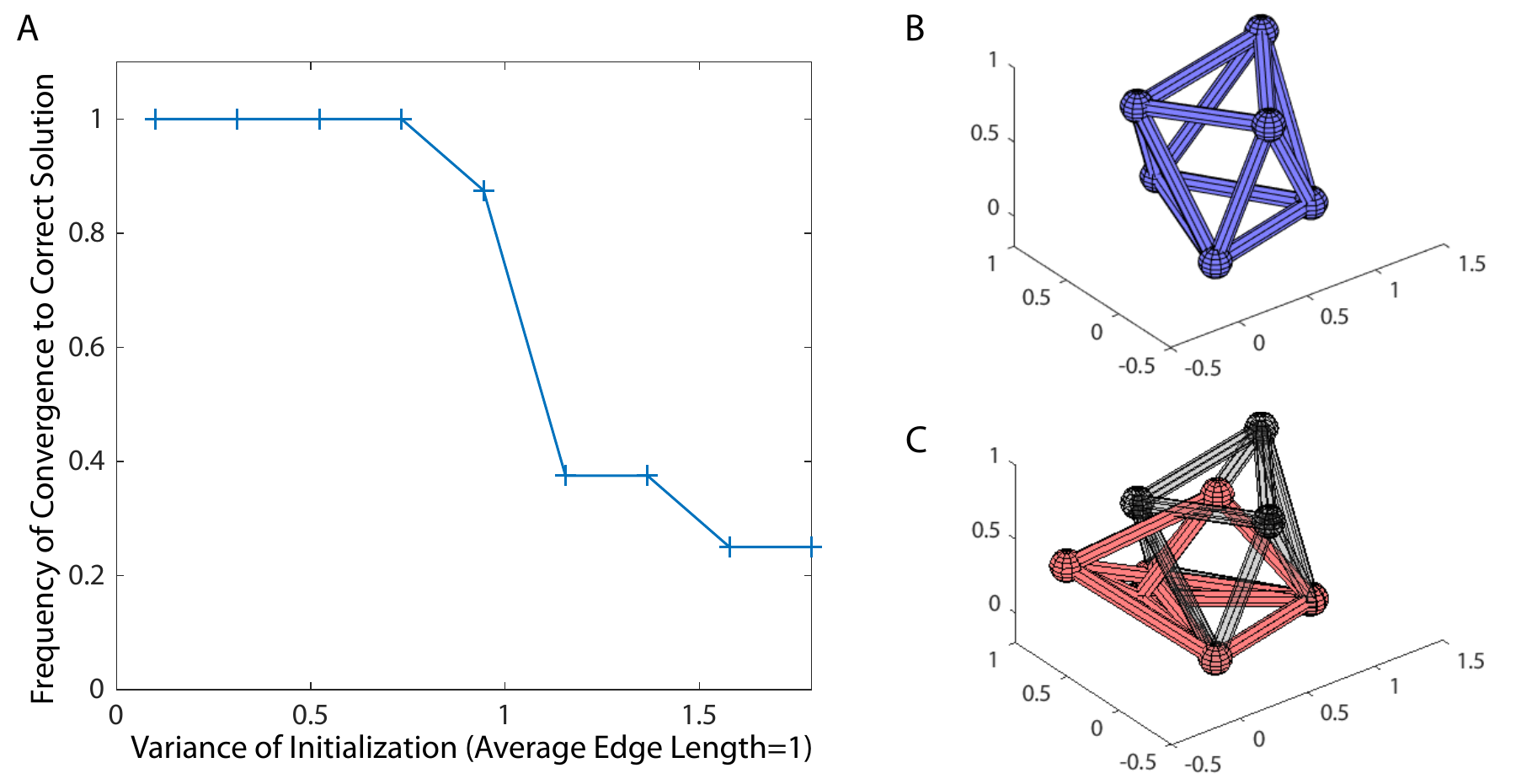}
\caption[Effect of Initial Guess on Convergence]{ (A) The rate of convergence to the solution for the distance based state estimation when different initializations of the octahedron are used. For small levels of variance, the optimization using relative distances converges to the solution. As variance increases, the results begin to converge to other solutions. (B) The correct configuration. (C) An incorrect configuration, which is a local minimum to the relative distance cost function, (Eq.~\ref{eq:D_J_measure}).   }
\label{fig:D_Initialization}

\end{figure}

One challenge of state estimation with relative distance measurements is the possible existence of multiple solutions.  For the case of relative position measurements, there is a unique minimizer to the objective function that satisfies the constraints. However, for the case of relative distance measurements, there are potentially other configurations that also locally minimize the cost function. To examine this effect, we repeat the previous experiments, but instead of adding noise to the measurements, we add noise to the initial guess. We added normally distributed noise with 0 mean, and ran the optimization at different levels of variance ranging from 0.1 m to 1.6 m. We completed 15 trials at each level of variance and determined the percentage of time the results converge to the true solution (Fig.~\ref{fig:D_Initialization}A). For noise with low variance, the result converged to the correct solution in all of the trial runs. With increasing variance of the noise added to the initial guess, the optimization will often converge to solutions different than the true configuration of the robot. The true configuration is shown in Fig.~\ref{fig:D_Initialization}B. Fig.~\ref{fig:D_Initialization}C shows an alternate configuration overlaid with the true configuration. In both of these configurations, all of the relative distances are identical. The amount of variance that can be introduced that results in convergence to the proper solution depends on the configuration of the graph, and may be lower or higher based on how the nodes are positioned.  From these results, we note that it is preferable, both in terms of convergence properties and computational speed, to be able to obtain relative position measurements. However, relative position measurements do require more sensing information than the relative distance measurements. 

\subsection{Distributed Control}
\begin{figure}[tp] 
\centering \includegraphics[width=\columnwidth]{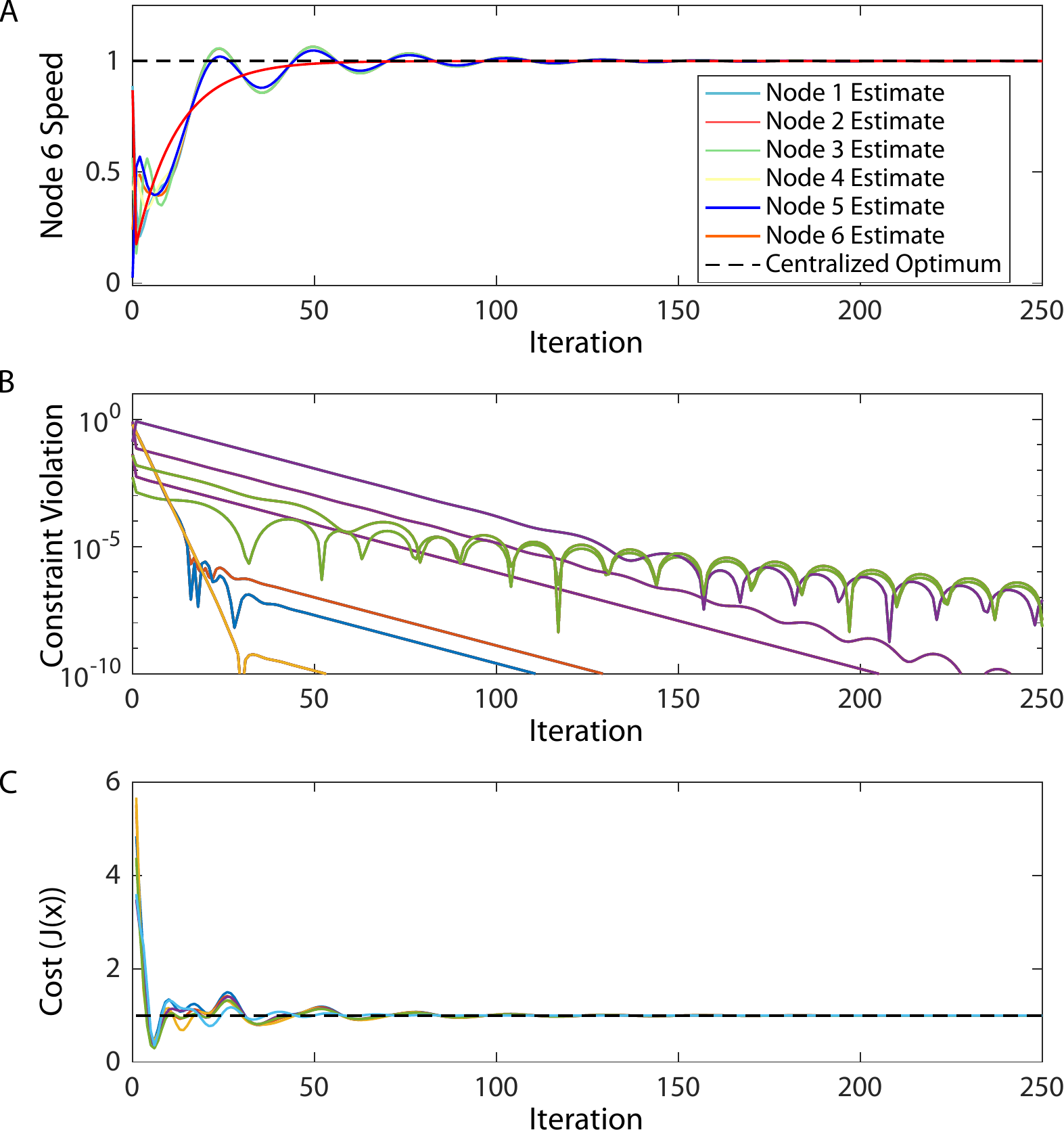}
\caption[Convergence Plots for Distributed Control]{Convergence of each node's local copy of all node's control commands. (A) The local copy maintained by each node of the commanded velocity of node 6. All local copies converge to an identical value. (B) The constraint violation as computed by each node. (C) The centralized cost evaluated based on each agent's local copy of the control commands. All agents converge to the same solution, which is identical to the solution obtained by solving the centralized optimization. }
\label{fig:D_Convergence}

\end{figure}

\begin{figure}[tb] 
\centering \includegraphics[width=\columnwidth]{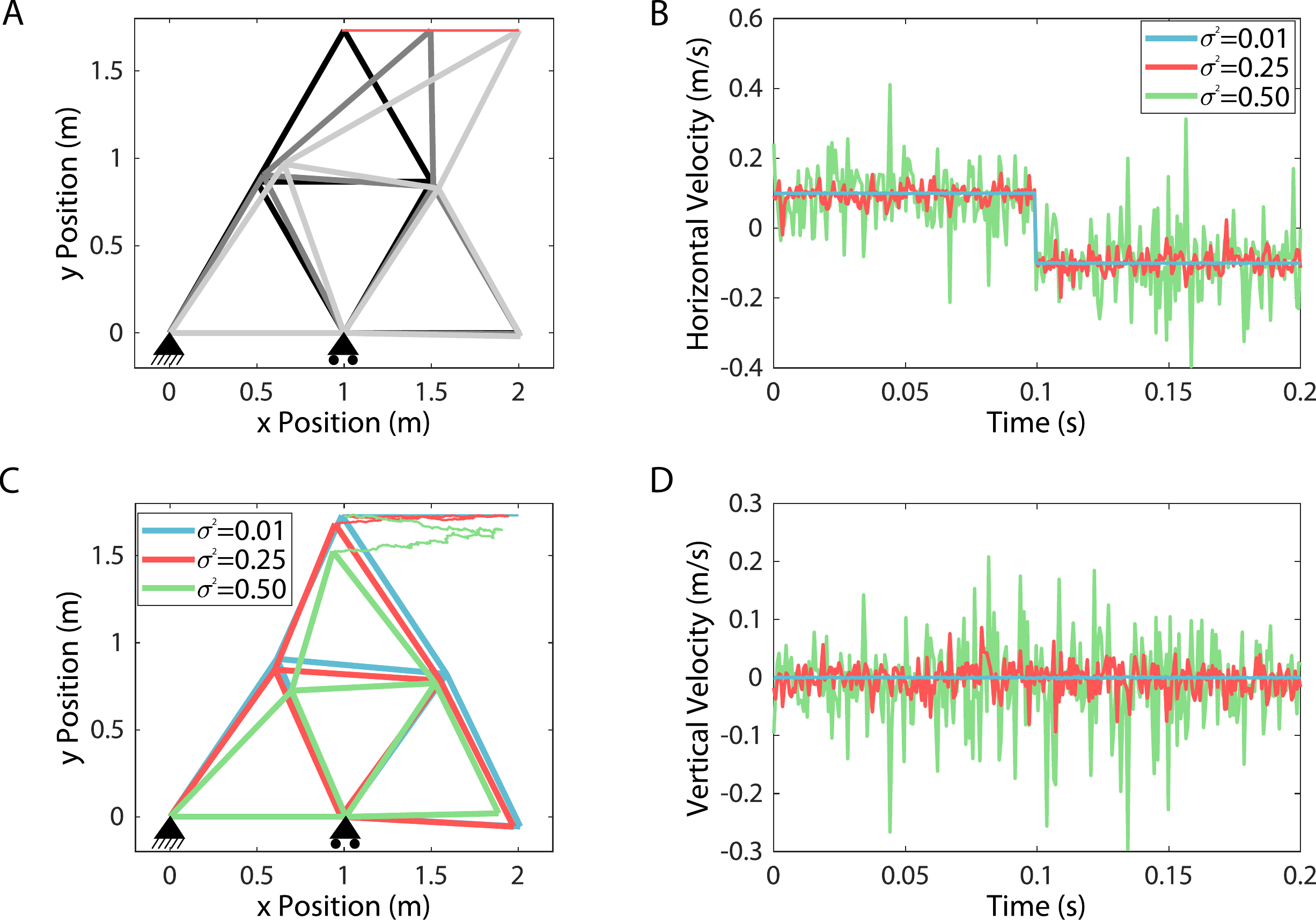}
\caption[Integrated Estimation and Control]{Integration of estimation and control algorithms for a 2D truss robot that consists of 9 actuators and 6 nodes. (A) Snapshots of the robot as it moves the top node with the prescribed velocity. (B) and (D) The measured velocity of the top node with different amounts of noise injected into the relative position measurements used in the estimation algorithm.  (C) The trajectories of the top node with the different amounts of noise, as well as the final configuration of the robot. With increasing measurement noise, the trajectory becomes less accurate.       }
\label{fig:D_Integrated_Est_Cont}
\end{figure}

In this section, we simulate the distributed control algorithm that uses the robot state as a starting point, and determines the velocities with which all nodes move. We consider a 2D truss robot consisting of 9 actuators and 6 nodes as shown in Fig.~\ref{fig:D_Integrated_Est_Cont}A. We assign the initial task of moving the top node (node 6) to move with a velocity of 1 m/s in the $x$ direction, while the cost function for the optimization is Eq.~\ref{eq:D_form_pos}.  We show the evolution of each node's local copy of all nodes' planned velocities during one round of ADMM updates in Fig.~\ref{fig:D_Convergence}. The results demonstrate that as all agents converge to an identical set of control commands, the constraint violation decreases, and the cost function converges to the minimizer of the centralized problem. The solution to the centralized problem is found by solving the optimization in Eqs.~\ref{eq:D_cent_J}-\ref{eq:D_cent_Axb} with identical cost function and constraints. 

We now evaluate the performance of the control scheme over multiple rounds of ADMM updates and when the controller is used in conjunction with the state estimation scheme described in Sec.~\ref{Sec:D_StateEstimation}. Fig.~\ref{fig:D_Integrated_Est_Cont} shows the performance of the robot when the task is to move the top node with a velocity of 1 m/s in the $x$ direction for 2 seconds, and then reverse the velocity. Each node measures the relative position of their neighboring nodes, and then reconstructs the state. For the control, we use the cost function in Eq.~\ref{eq:D_cost_quad}, fix the position of the bottom left node in both the horizontal and vertical directions, and the position of the bottom center node in the vertical direction as illustrated in Fig.~\ref{fig:D_Integrated_Est_Cont}A. This task is performed open-loop, with the command being to always move with a constant velocity in the $x$ direction, regardless of whether past errors have occurred. The behavior of the robot without noise is shown in Fig.~\ref{fig:D_Integrated_Est_Cont}A. 

To evaluate the behavior in a realistic scenario, we add noise to the measurements each node takes of the relative position of its neighboring nodes. We use these noisy measurements to compute the estimated robot state and  the control action, and then we apply the control to the actual state of the robot. The noise is normally distributed with $0$ mean and variances of $0.01$ m, $0.25$ m, and $0.50$ m. The trajectories of the top node, as well as the final configuration of the robot, are shown in Fig.~\ref{fig:D_Integrated_Est_Cont}C. The velocity of the top node is also shown in Fig.~\ref{fig:D_Integrated_Est_Cont}(B and D). With increasing noise, the state estimate of the robot becomes less accurate, as expected, resulting in rapid changes in the commanded velocity, including changes in the vertical velocity despite the fact that no change in vertical velocity is desired. However, despite this noise, the overall motion of the robot is still as desired.  These results demonstrate that even in the case of noise with a variance in measurements that are $\frac{1}{4}$ or $\frac{1}{2}$ of the nominal edge length, the algorithm leads to behaviors that are similar to the nominal behavior without noise. This indicates that the integration of the state estimation and control algorithms is robust to noisy measurements that may be encountered in real-world situations. A video of this experiment is included in the supplementary materials. 

\section{Distributed control of an Isoperimetric Robot}
A particular type of truss robot that has shown promising capabilities is the isoperimetric robot presented in \cite{usevitch2020untethered}. This robot consists of a set of inflated fabric tubes and a set of robotic roller modules. The roller modules pinch the tube between sets of cylindrical rollers while still allowing airflow, creating a region of reduced bending stiffness in the inflated beam that acts as an effective joint. These roller modules can then be connected together through universal joints, creating a truss structure where each inflated tube makes up multiple edges, and the nodes of the structure consist of robotic roller modules. The robot changes shape not by lengthening and shortening individual edges, but by driving the roller modules along the tube by spinning the cylindrical rollers, simultaneously lengthening one edge and shortening another. This means that the total length of the tube remains constant, leading to the term isoperimetric, or constant perimeter. Since the total edge length remains constant, the total internal volume of the inflated tubes is approximately constant, meaning that this large, inflated robot can operate without an active air source.  This isoperimetric constraint is an additional constraint that is not present in conventional truss robots. In this section, we discuss the application of distributed estimation and control techniques to isoperimetric truss robots. 

\begin{figure}[tb] 
\centering \includegraphics[width=\columnwidth]{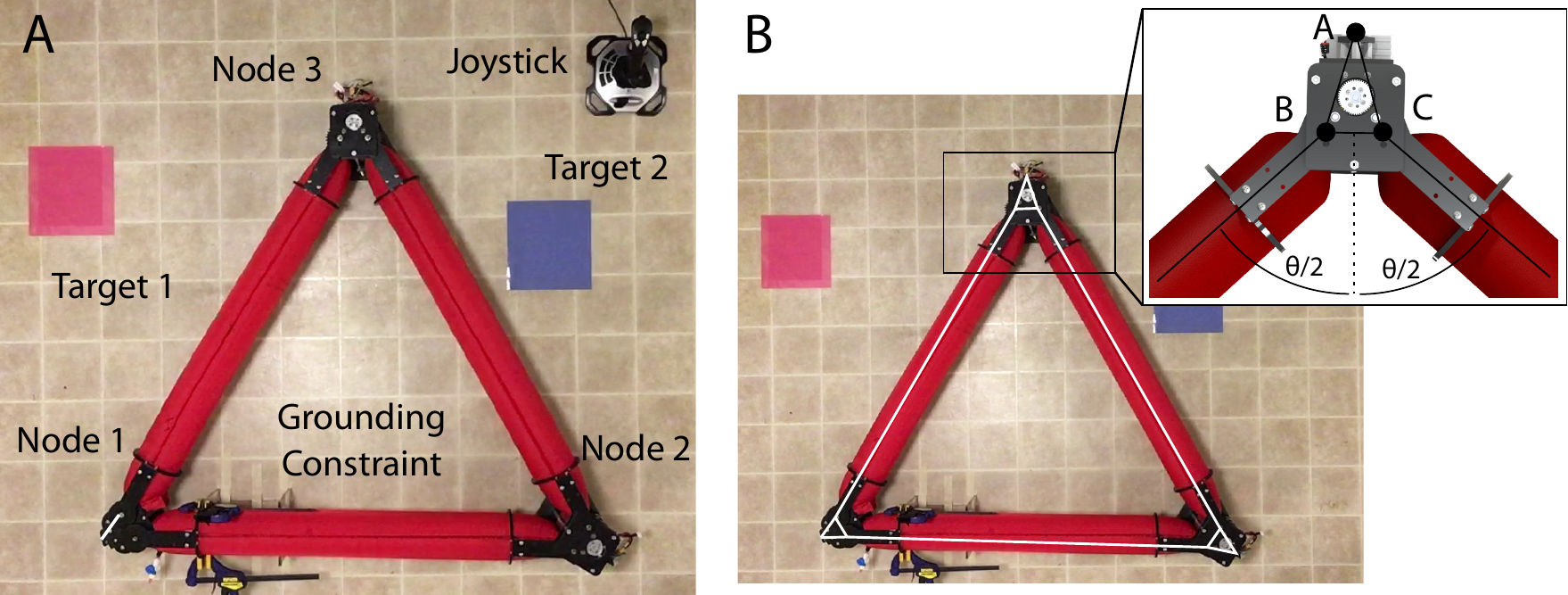}
\caption[Effect of Initial Guess on Convergence]{Isoperimetric robot used for this experiment. (A) shows the three nodes of the robot, including a passive node that constraints the ends of the tube (Node 1) and two motorized nodes (node 2 and 3).  Two target regions are also shown. (B) Illustration of the kinematics of an individual node. The physical extent of the device requires that the state of each node is represented by three points. A physical constraint ensures that a line from the midpoint of the two rollers to the top point bisects the angle formed by the adjacent edges of the tube. }
\label{fig:D_IsoP_Setup}

\end{figure}

For the experiments in this paper, we use a single triangle isoperimetric truss robot as shown in Fig.~\ref{fig:D_IsoP_Setup}. The triangle consists of one passive node (Node 1) that holds the ends of the inflated tube.  The two other nodes are motorized, and are capable of moving along the tube under the power of an electric motor. Each rollers tracks the distance is has traveled along the inflated tube using an encoder. The distance along the tube maintained by each roller allows node 3 to know the length of the node between nodes 1 and 3, and node 2 to know the length of the edge between node 1 and node 2.  The length of the edge between node 2 and node 3 can be inferred using knowledge of the constant total tube length and the lengths of the two other edges. 

\subsection{Additional Constraints on the Isoperimetric Robot}

The kinematics of the isoperimetric robot differ from other instantiations of the truss robot in that the robot can control and sense the distance that each roller has moved along the tube, as opposed to the overall edge lengths. We denote the distances along the tube from the beginning of the tube to each roller as a vector $r$. The relationship between the distance of the rollers from the beginning of the tube to the edge lengths is

\begin{equation}
    L=B r + \begin{bmatrix} 0 \\ L_{tot} \end{bmatrix}
\end{equation}

Where each row of the $B$ matrix consists of all zeros except for a one in the column corresponding to the beginning of the edge, and a $-1$ in the column corresponding to the end of each edge. While it is possible for all of the nodes to move along the tube, we choose to have one node stationary, which houses the beginning and end of the tube. 

As previously stated, we perform the distributed optimization in terms of the node positions. When the nodes converge on the proper estimate of the total motion, $\dot{x}^*$, and each roller extracts the speed at which it must move along the tube as follows: 

\begin{equation}
    B^T R(x) \dot{x}= \dot{r}
\end{equation}

In addition, we have a constraint that the total length of the tube remains constants, which can be expressed as
\begin{equation}
1^T L(x) = L_{tot}
    \end{equation}
Taking the derivative, we obtain the following relationship between the speed of the nodes
\begin{equation}
    1^T\dot{L}=1^T R(x) \dot{x} = 0
\end{equation}

Applying this constraint in a distributed fashion requires that each node has knowledge of the total length of the tube $L_{tot}$.

\subsubsection{Constraints in the Physical Implementation of the Robot}
 The physical construction of the isoperimetric truss robot requires some modification to the kinematic model of the robot presented in Sec.~\ref{Sec:Formulation}. A close up of a single roller module and an illustration of its kinematics are shown in Fig.~\ref{fig:D_IsoP_Setup}B. Due to the physical extent of the inflated tube, two separate roller mechanisms pinch the tube in order to allow the tube to bend to a small angle without the tube experiencing self interference. We represent the state of each roller module as the position of three points as shown in Fig.~\ref{fig:D_IsoP_Setup}B: point A is located at the top of the module and points B and C are located at the positions where the tube is pinched by the two sets of rollers. In a larger truss structure, point A would correspond to the connection point between adjacent nodes, and the location of all of these joints would comprise the state of the robot. In the case of the isoperimetric robot, the state of the robot is represented by the location of all three of these points (points A, B and C).
 
 To account for the increased size of the robots state, several physical constraints are included in the construction of the robot that ensure that the robot is fully defined. The physical construction of the roller module ensures the distance between points A, B and C is fixed. This is expressed mathematically as  
 \begin{equation}
     \|p_{iA}-p_{iB}\|= \|p_{iA}-p_{iC}\|=L_{AB}
 \end{equation}
  \begin{equation}
     \|p_{iB}-p_{iC}\|= L_{BC}
 \end{equation}
 where $L_{AB}$ is the distance between the top point and the rollers, and $L_{BC}$. 
We denote the constraint edge length $L_{con, i}$, and an estimate of the overall robot shape must satisfy
\begin{equation}
\|L_{con, i} - L_{con,i}(x)\|^2=0. \label{eq:Extra_Lengths}
\end{equation}

An additional constraint that is mechanically included into the design of the roller modules is that point A always lies on a line that bisects the two incoming edges. Mechanically, this constraint is enforced by a gearing that connects the two angled arms. Mathematically, this constraint is expressed as: 

\begin{equation}
    \frac{(x_d-x_b)^T(x_c-x_b)}{\|x_d-x_b\| \|x_c-x_b\|}=\frac{(x_e-x_c)^T(x_b-x_c)}{\|x_e-x_c\| \|x_b-x_c\|}. \label{eq:Bisection}
\end{equation}

We note that in three dimensions, an additional constraint also exists that each of these sets of nodes remain in plane.  As we are dealing with a single triangle in the plane in this experiment, we need not include this constraint in our calculations.
 
For notational convenience, we define the constraint of the additional edge lengths as discussed in \ref{eq:Extra_Lengths}, and the bisection constraint expressed in \ref{eq:Bisection} at the $i$th node as

\begin{equation}
    Q_{con,i}(x)= 0.   \label{eq:Bisection}
\end{equation}
By taking the derivative of this expression with respect to time, we obtain an expression linear in the velocity of the nodes that constrains the possible motions of the device while respecting these constraints
\begin{equation}
    \frac{dQ_{con,i}(x)}{dx} \dot{x}= 0.   \label{eq:dQdt}
\end{equation}


\subsection{Distributed Control Sketch}
In the previous section we presented the additional constraints that must be added to the basic truss robot framework when considering the isoperimetric robot. In the general case of an isoperimetric robot this includes the constraint that the total edge length remains constant. In the specific case of the physical embodiment of the isoperimetric robot that we use in this experiment, this includes the addition of more kinematic points to the state of the robot, and more physical constraints that define the motion of these points.  In general, we note that the increased size of the total state increases the computational difficulty of this problem.  We now summarize how these affects impact the algorithms that the robot uses to compute its distributed control actions.

For the physical isoperimetric robot, the cost function for the distributed optimization used for the state estimation problem 
 
 \begin{equation}
     \|L_{tot}- \sum L(x) \|^2 + \sum_{i=1}^{N_L} \|-L_i(x)-L_{m,i}\|^2  +\sum_{i=1}^n Q_{con, i}(x). 
 \end{equation}

This equation serves as the equivalent to Eq.~\ref{eq:D_J_measure}, with the inclusion of the additional constraints that deal with the constant perimeter and the added kinematic constraints. 

The control problem is executed in a similar manner to the algorithm shown in Eqs.~\label{eq:D_vel_con}-eq.~\label{Motion_Primitive}, but with the additional constraint that the total edge length remains constant, and that the additional constraints are enforced

\begin{align}  
   &\min_{\dot{x}_i}    
   \begin{aligned}[t]
        J_i(\dot{x}_i)  \label{Motion_Primitive}
   \end{aligned} \\
   &\text{subject to} \notag \\
   &\quad \begin{bmatrix} A_i \\ C_i \\ \frac{dQ_i(x)}{dx} \\ 1^T R(x) \end{bmatrix} \dot{x}_i=\begin{bmatrix} b_i \\ 0 \\0 \\0 \end{bmatrix}  \label{eq:lin_control} \\
   &\quad  \dot{x}_i=\dot{x}_j, \quad \forall j \in N_i,   \label{eq:D_vel_con} 
\end{align}



The implementation of this controller is equivalent to the methods presented for the standard truss robot.Communicating only with their neighboring nodes, the nodes use the distributed ADMM framework to update their individual estimate of the global state. They then use the same framework to iteratively solve the control problem, and converge on the desired motion of each node. While the information measured by the nodes is different and the control action is different for the isoperimetric robot, the information exchanged between the nodes is identical. This shows a level of flexibility of the algorithm for including different types of sensors and actuators in the framework. 


\subsection{Demonstration of Control with the Isoperimetric Robot}

To validate the performance of the algorithm we perform tests using the distributed algorithm with robotic hardware. Each node is equipped with a Teensy microcontroller that runs a proportional-integral-derivative controller that moves the roller module along the tube with a target speed. During the experiment, the two Teensy 3.2 microcontrollers measure the nodes' position along the tube via incremental encoders and broadcast their positions to an offboard laptop over radio. All estimation and control computations are performed in a distributed manner on the offboard computer, and the desired velocity along the tube is broadcasted to each roller module. In this demonstration, a user teleoperates the Node 3 (Fig.~\ref{fig:D_IsoP_Setup} of the robot using a joystick. 

\begin{figure}[tb] 
\centering \includegraphics[width=\columnwidth]{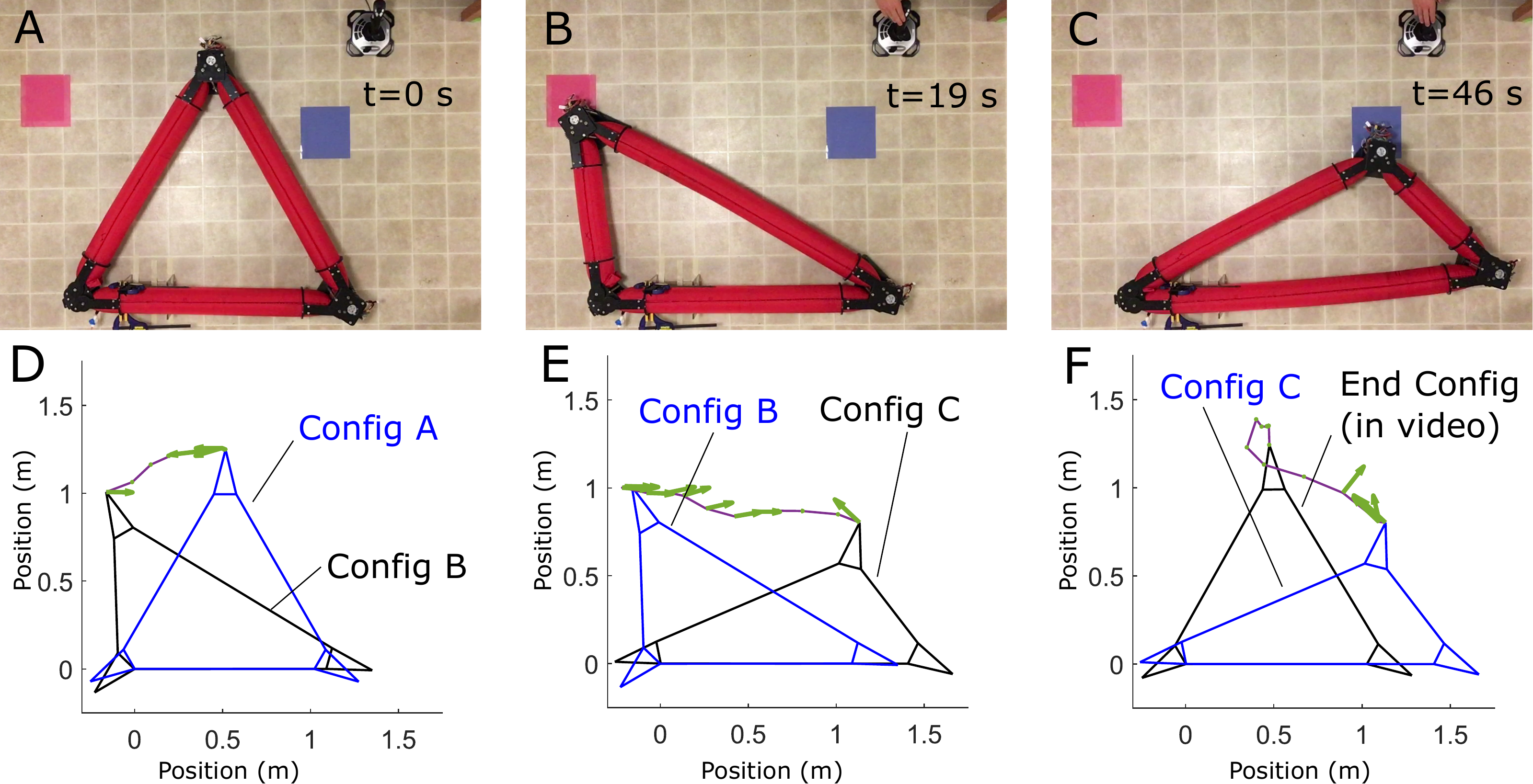}
\caption[Demonstration]{ Demonstration of the isoperimetric robot moving its top node between targets (shows as the blue and red squares) while running the distributed estimation and control algorithm. A user teleoperates the top node of the robot. The desired motion is communicated only to the top node, but through the distributed algorithm the nodes of the robot work together to achieve the desired motions. D,E, and F show the robot's estimate of its state, and the arrows at the top node show the commanded velocity from the teleoperator.}
\label{fig:Ex_Results}

\end{figure}

The results of the demonstration are shown in Fig.~\ref{fig:Ex_Results}. Using input from a joystick, a user is able to drive the top node of the robot from its initial position (Fig.~\ref{fig:Ex_Results}A) to a target to the left (Fig.~\ref{fig:Ex_Results}B), to a target to the right (Fig.~\ref{fig:Ex_Results}C). Fig.~\ref{fig:Ex_Results} D,E,F show the robots' estimate of its own state, as well as each node's local copy of the commanded velocity of the top node, shown as green arrows.  A video of this experiment is included in the supplementary materials. 

During these experiments, significant latency is introduced due to the dependencies on the communication between the base computer and the nodes.  This latency could be substantially reduced by performing computation on the nodes themselves, improved communication protocols, and better software engineering.  However, despite this significant latency, a user is still able to teleoperate the top node of the robot to reach two target configurations.  Each node in the robot, in a distributed manner, estimates the positions of all nodes and then coordinates its control actions to achieve the specified task-space behavior.



%

\section{Conclusion}
This paper presented distributed control algorithms for truss robots composed of extensible links connected at universal joints, and then demonstrated these algorithms in simulations and in an experiment with an isoperimetric truss robot. These algorithms, based on a consensus ADMM framework, allow the nodes to coordinate their behavior across the entire network while only communicating locally with their physical neighbors.  Each agent uses an iterative ADMM update to reconstruct the state of the entire robot during the state estimation phase while exchanging messages that are each agent's estimate of the global state. During the control phase, the ADMM updates are performed on the local commanded velocities of each node. In the case of a quadratic cost function, the updates are extremely efficient computationally.  However, they do require many rounds of communication, (for the examples given in this paper, on the order of 100s of communication rounds).   

The control approach presented in this paper coordinates motions of the actuators of a truss robot to achieve a set of desired node motions, but these methods do not determine what the desired node motions should be. A separate high-level planner could be developed to provide the desired motions. The approach of separating the motion specification and the coordination of the actuators motion can potentially simplify the planning problem for the high-level planner, as it would only need to plan for the motion of a subset of the robot's nodes. The desired motions could also be provided from a human operator, who could teleoperate one node or the center of mass of the robot using a joystick. The desired motions could also be generated based on sensor information obtained locally at each node.  For example, if a node observes a target with an onboard camera, it could determine to move the center of mass in the direction of the target. Future work will examine these different types of high-level planners. 

In the current system, each agent reasons about the position and desired velocity of all other nodes in the network. This means that the size of the messages passed between neighbors grows linearly with the number of agents in the network, even if the number of neighbors stays the same. Also, this system requires that all agents have aligned frames. Some of these requirements could be relaxed to increase the efficiency, scalability, and practicality of these distributed algorithms in the future. 


\bibliographystyle{IEEEtran}
\bibliography{IEEEabrv,bibliography}

\end{document}